\documentclass[runningheads]{llncs}
\usepackage[misc,geometry]{ifsym}

\usepackage{graphicx}
\usepackage[dvipsnames]{xcolor}
\usepackage{multirow,tabularx}
\usepackage{comment}
\newcolumntype{Y}{>{\centering\arraybackslash}X}

\renewcommand{\tabcolsep}{5pt} 
\begin{document}
\title{HealNet - Self-Supervised Acute Wound Heal-Stage Classification}
%
%
\author{Héctor Carrión$^1$, Mohammad Jafari$^3$, Hsin-Ya Yang$^2$, Roslyn Rivkah Isseroff$^2$, Marco Rolandi$^1$, Marcella Gomez$^1$, Narges Norouzi$^{1*}$}
\authorrunning{Carrión et al.}
%
\institute{University of California, Santa Cruz$^1$, University of California, Davis$^2$, \\ Columbus State University$^3$ \\
\email{nanorouz@ucsc.edu$^*$}} 
%
\maketitle              
\begin{abstract}
Identifying, tracking, and predicting wound heal-stage progression is a fundamental task towards proper diagnosis, effective treatment, facilitating healing, and reducing pain. Traditionally, a medical expert might observe a wound to determine the current healing state and recommend treatment. However, sourcing experts who can produce such a diagnosis solely from visual indicators can be difficult, time-consuming and expensive. In addition, lesions may take several weeks to undergo the healing process, demanding resources to monitor and diagnose continually. Automating this task can be challenging; datasets that follow wound progression from onset to maturation are small, rare, and often collected without computer vision in mind. To tackle these challenges, we introduce a self-supervised learning scheme composed of (a) learning embeddings of wound's temporal dynamics, (b) clustering for automatic stage discovery, and (c) fine-tuned classification. The proposed self-supervised and flexible learning framework is biologically inspired and trained on a small dataset with zero human labeling. The HealNet framework achieved high pre-text and downstream classification accuracy; when evaluated on held-out test data, HealNet achieved 97.7\% pre-text accuracy and 90.62\% heal-stage classification accuracy. 
\keywords{Wound healing  \and Self-supervised medical image processing \and Clustering.}
\end{abstract}
\section{Introduction and Related Work}
Wound healing is a variable, painful, and often prolonged process that can impact the quality of life for millions of people worldwide. Approximately 6.5 million patients present chronic wounds annually in the United States, requiring a treatment cost of over \$25 billion per year \cite{sen2009human}. In 2014, acute wounds resulted in 17.2 million hospital visits \cite{steiner2020surgeries}, while the odds of an acute wound requiring further care significantly increases with age, diabetes, and obesity \cite{sen2009human}; thus a large portion of acute wounds will not resolve uneventfully and burden healthcare resources. An accessible and automated protocol to identify, track, and predict wound heal-stages could help tackle these challenges while reducing costs.

Biologically routine wound healing mechanics are four precise and programmed stages: hemostasis, inflammation, proliferation, and maturation \cite{guo2010factors}. Numerous chemical reactions and interactions between cells occur within these phases under and above the wound bed. However, the visual indicators that result from or relate to each stage are medically under-specified. Therefore, dermatologists are often limited to studying wound area changes or rely on mice wound re-epithelialization and histological analysis (which requires harvesting the wound-bed and thus sacrificing the mice) to understand wound-stage progression \cite{bagood2021re}. Invasiveness, the cost of long-running experiments, and the lack of qualified experts that can analyze image-based results contribute to the lack of large-scale, complete wound-stage progression image datasets.

\begin{figure}[t]
    \centering
    \includegraphics[width=0.75\textwidth]{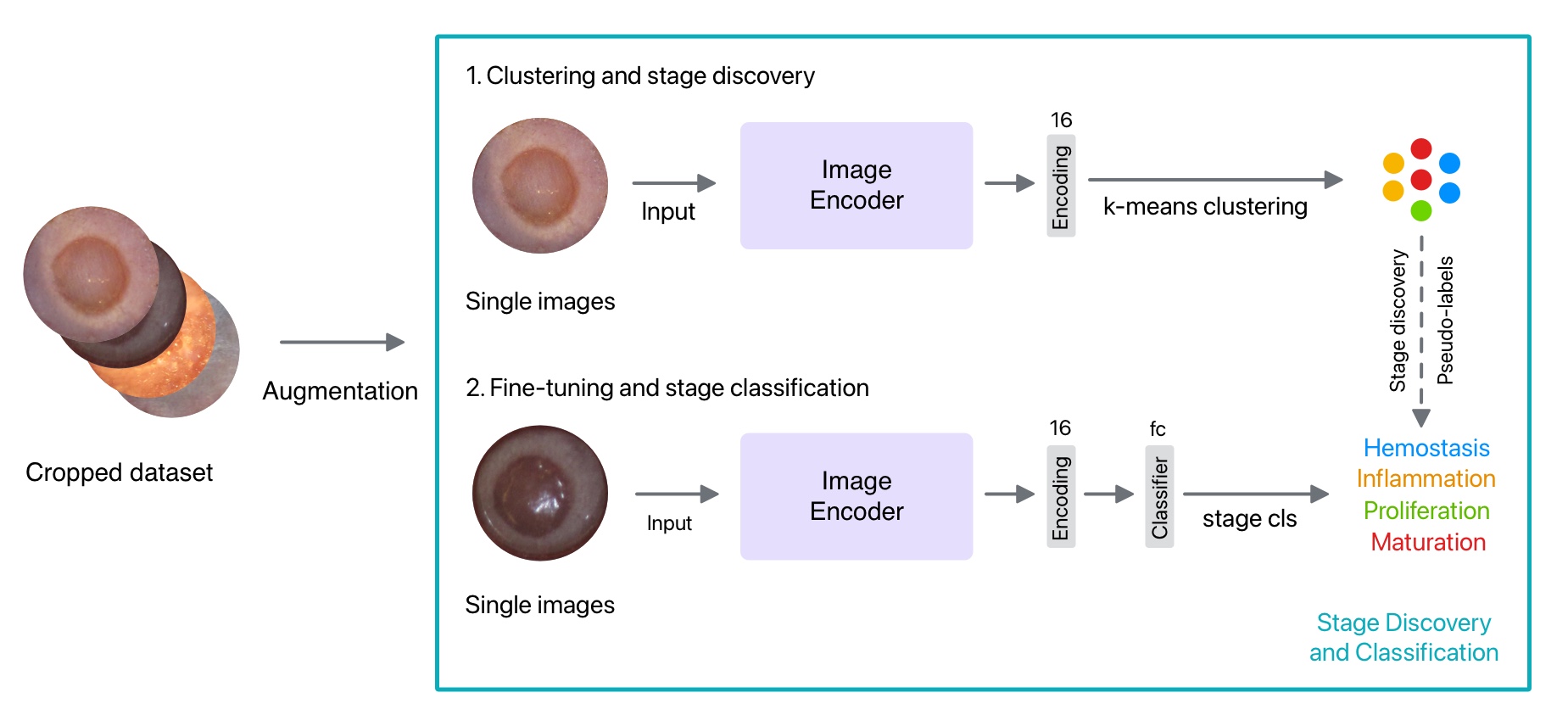}
    \caption{\textbf{(Step 1)} The self-supervised network is used to generate 16-dimensional feature vectors. We cluster these feature vectors using k-means with $k=4$. \textbf{(Step 2)} The new pseudo-labels (based on clusters) are used to assign each image a heal-stage class. The network is then re-tasked and fine-tuned using pseudo-labels, predicting the wound stage.}
    \label{fig:clustering}
\end{figure}

Machine Learning (ML) for automated wound analysis has not been approached from the heal-stage angle. Methods exist for wound detection and size estimation \cite{chino2020segmenting}, where obtaining ground-truth labels are relatively straightforward. Other classification systems aim to identify different characteristics of wounds or ulcers \cite{oyibo2001comparison}. Acute wound linear regression models incorporate location, age, size, and infection features to generate time-to-heal estimates \cite{ubbink2015predicting}. Recent lesion segmentation work has found significant inter-and-intra annotator variance even among experts \cite{mirikharaji2021d}, suggesting that precisely defining wound-stage visual indicators is not trivial.

Weakly-supervised and self-supervised learning approaches have shown the ability to improve the performance of tasks with little or no annotated data by optimizing a pre-text objective function that learns representations without (or with few) explicit human labels. For example, learning color from grayscale images (colorization) \cite{larsson2016learning,zhang2016colorful}, shuffling and predicting image tile permutations (Jigsaw) \cite{noroozi2016unsupervised}, removing and re-generating image regions (inpainting) \cite{pathak2016context,ledig2017photo} rotation prediction (RotNet) \cite{gidaris2018unsupervised} and classifying augmented positive and negative pairs (contrastive learning) \cite{chen2020improved,chen2020simple}. In modalities like video and audio, the inherent temporal structure of the data can be exploited to create other proxy objective functions, for example shuffling video frames and predicting temporal order \cite{misra2016shuffle}. Further work has shown that these embeddings can be clustered and re-used as pseudo-labels to improve representation quality for weakly-supervised and self-supervised models \cite{yan2020clusterfit}.

Due to the nature of wound healing, capturing progression over time involves learning the inherent temporal structure. Thus, a proxy task is created to learn wound healing temporal dynamics. We study a small, unlabeled dataset containing daily mouse wound observations from onset to maturation. We leverage the dataset to generate wound image pairs in the temporal forward (positive) and backward (negative) directions. This step increases our dataset size by a factor of $15$. The wound pairs are then used to train a network to determine the input pair's temporal validity and thus encoding features relevant to wound healing temporal dynamics. These embeddings are then clustered into four groups (one per heal-stage) to generate pseudo-labels without human intervention. Finally, a classifier is re-tasked and fine-tuned on pseudo-labels to predict the heal-stage of input images (Fig. \ref{fig:clustering}).

\begin{figure*}[t!]
    \centering
    \includegraphics[width=0.75\textwidth]{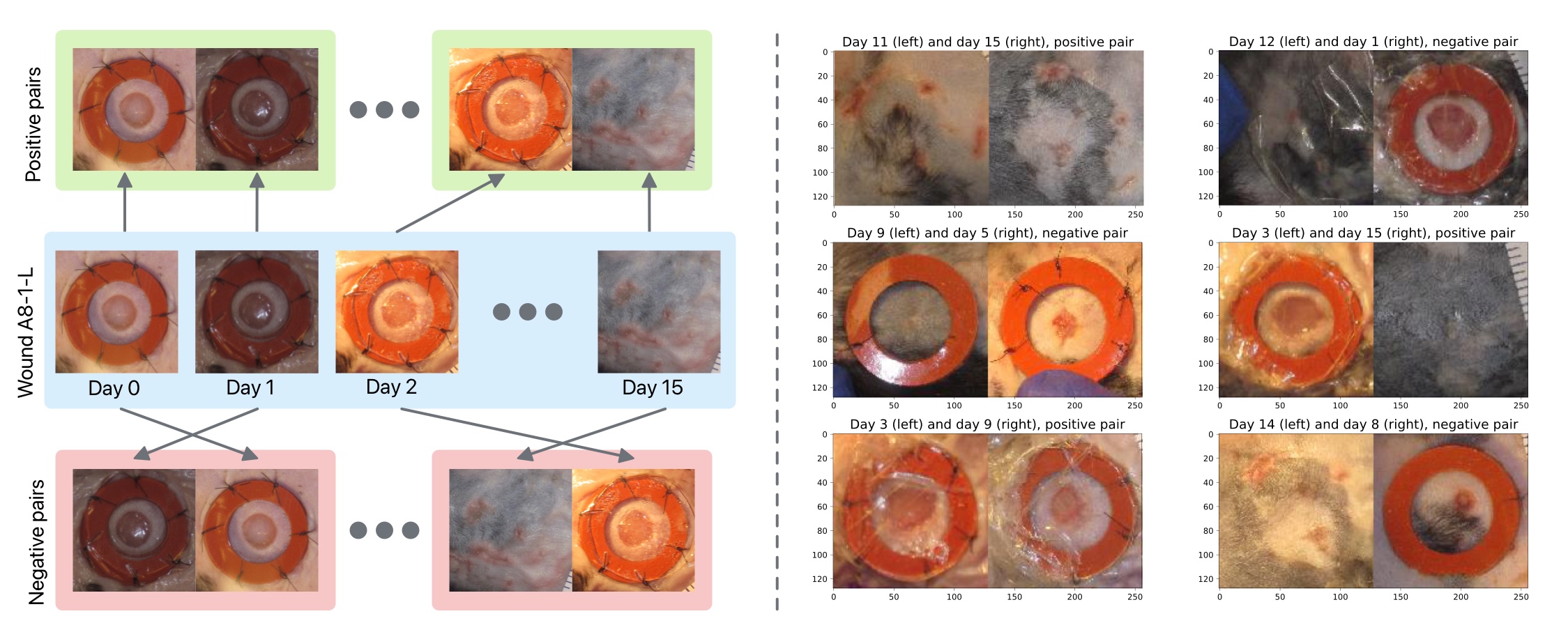}
    \caption{\textbf{(left side)} Due to the nature of wound healing, images capturing progression naturally contain a temporal structure. We generate image pairs in the forwards (positive) and backward (negative) temporal directions; this serves as the input for our self-supervised temporal coherency training step. \textbf{(right side)} Randomly sampled image input pairs.
}
    \label{fig:data}
\end{figure*}

\section{Description of Data}
The experiment for generating the image dataset used in this work is described in \cite{carrion2021automatic}, from which we generate circular wound-only crops. The images contain wounds from eight mice (four young and four aged) imaged daily over sixteen days. Each individual had a wound inflicted on the left side and another on the right side. The resulting dataset is 256 images (8 mice $\times$ 2 wounds $\times$ 16 days). Note that this dataset was not captured with computer vision in mind; consequently, we need to account for challenges such as blur, occlusion and illumination noise. To address this, data augmentation was used in training.

Wound pairs have been created for the pre-text learning task in order to capture the temporal healing process. We generated pairs in the temporal forwards (positive) and backwards (negative) directions. Fig. \ref{fig:data} illustrates the process of creating these pairs. We did not generate pairs containing same-day same-wound images. Thus, the resulting dataset contains 3,840 image pairs (16 wounds $\times$ 16 days $\times$ 15 pairs per day). In order to allow our validation and test sets to capture any differing dynamics between the two cohorts of young and aged mice, we randomly selected two young wounds and two aged wounds from the dataset; a couple of unique aged and young wounds is placed into the validation set and a second couple into the testing set. After this split, the resulting image pair dataset contains 2,880 training samples (75\%), 480 validation, and testing samples (12.5\% each). The single image dataset then contains 192 training samples (75\%), 32 validation, and testing samples (12.5\% each). We consistently keep the same validation and test wounds held out throughout the training pipeline and between our pre-text and downstream tasks. Finally, we cropped identical circles around the centroid of each wound image to ensure no signals other than the wound itself are contained in the training data.

\begin{figure*}[t!]
    \centering
    \includegraphics[width=0.75\textwidth]{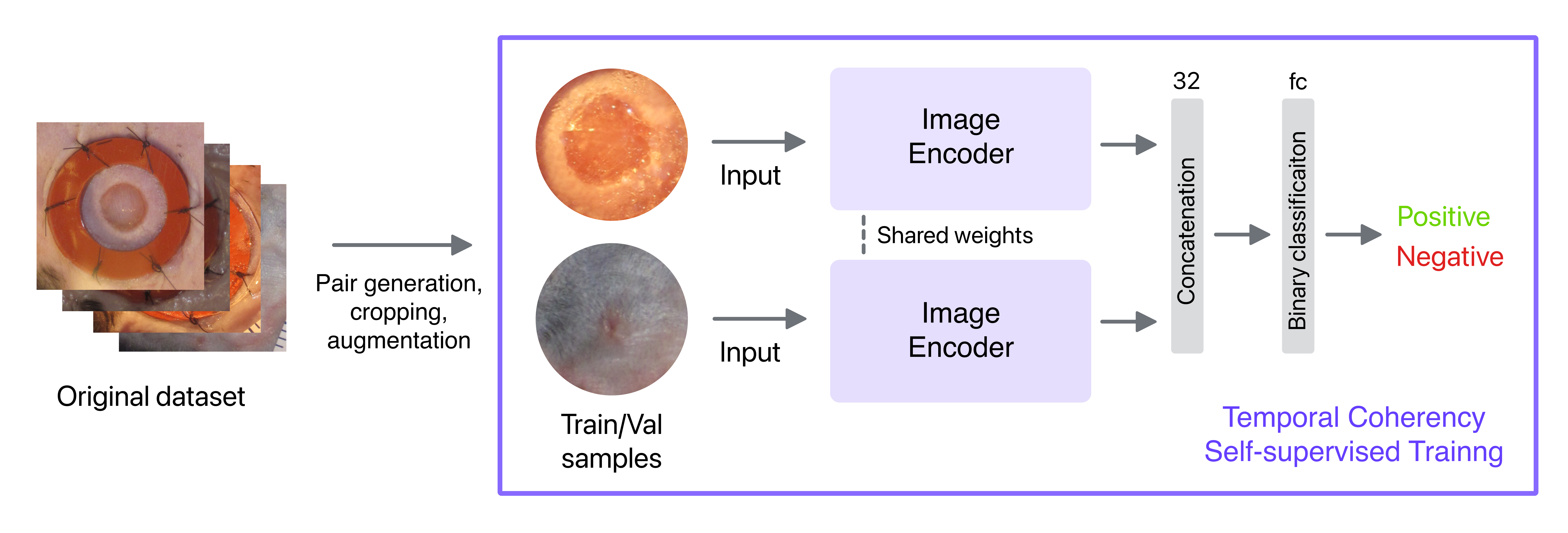}
    \caption{Positive and negative wound pairs train the network. The image encoder forms a Siamese tuple configuration with shared parameters; each branch outputs a 16-dimensional feature vector. The feature vectors are concatenated and passed through a fully-connected binary classifier, outputting the input pair temporal validity.}
    \label{fig:temporal-encoder}
\end{figure*}

\section{Approach}
The HealNet scheme involves a self-supervised learning step, followed by a clustering step, and a second learning step for wound-stage classification. Across all steps, the same image encoder neural network backbone is used and the trained weights are kept. Given the limited size of our training dataset, we chose DenseNet121 \cite{huang2017densely} as the HealNet backbone architecture. DenseNet is fully convolutional and thus includes fewer parameters than other commonly used feature extractors while showing strong performance metrics. The first learning step revolves around self-supervised pre-text learning and thus includes a concatenation layer and a fully-connected binary classifier as the network head. The second learning step resembles a more traditional image classifier, replacing the previous head with a single fully-connected layer for heal-stage classification. Before training, DenseNet is initialized to ImageNet  \cite{deng2009imagenet} weights.

\subsection{Temporal Coherency Self-supervised Training}
To predict the temporal validity of our image pairs and thus encode temporal dynamics, we define a tuple Siamese architecture \cite{bromley1993signature} using two parallel stacks of identical feature extraction networks. Each stack shares weights and takes a single image from the pair as input while outputting a representation encoding (of size 16). We concatenate both encodings and feed them into a single fully-connected binary classification layer that determines whether the input pair is negative or positive. The loss function is binary cross-entropy and the optimizer is Adam \cite{kingma2014adam} with default parameters. Fig. \ref{fig:temporal-encoder} summarizes the temporal coherency training scheme.

\subsection{Unsupervised Wound Stage Clustering}

In order to identify and cluster wound healing stages, we use unsupervised k-means clustering. We apply unsupervised k-means clustering to the representations generated by our temporal coherency network to generate our clusters. We set $k=4$ since medical literature suggests four main healing stages: hemostasis, inflammation, proliferation, and maturation. We use these cluster mappings to assign pseudo-labels to our dataset. Fig. \ref{fig:clustering} illustrates the two steps of assigning pseudo-labels using k-mean clustering and subsequently using these pseudo-labels for heal-stage classification \cite{yan2020clusterfit}. The question of how we specifically relate each cluster to each heal-stage is covered in Section \ref{sec:results}.

\subsection{Wound Stage Classification}

As shown in Fig. \ref{fig:clustering}, the final step of the pipeline is to re-task and fine-tune the pre-text network towards predicting the heal-stage of input images based on cluster assigned pseudo-labels. The re-tasking is done by adding a fully-connected layer with a softmax activation function and four output neurons (one per healing stage). The fine-tuning is accomplished by further training and updating the weights of the model using a low learning rate ($\eta =0.001$) and the Adam \cite{kingma2014adam} optimizer.

\section{Results and Discussion}
\label{sec:results}
This section covers loss and accuracy scores for both learning steps, as well as evaluates the quality of cluster-based stage discovery from a cluster-member appearance perspective, from a cluster-member wound-age statistics perspective and via agreement against human heal-stage annotators. Finally, we will cover the metrics achieved by the model on held-out test data at each learning step. 

\subsection{Temporal Coherency Self-supervised Evaluation}
The self-supervised model is trained until convergence. Our results show the network is not overfitting, achieves high accuracy on the validation set, and can predict the temporal validity of the train, validation, and test sets. The prediction accuracy for the temporal validity task is 97.2\%, 95.0\%, and 97.7\% on training, validation, and test sets, respectively, as compiled in Table \ref{tab:table2}. 

\subsection{Wound Healing Clustering Evaluation}
\begin{figure}[h!]
    \centering
    \includegraphics[width=0.45\textwidth]{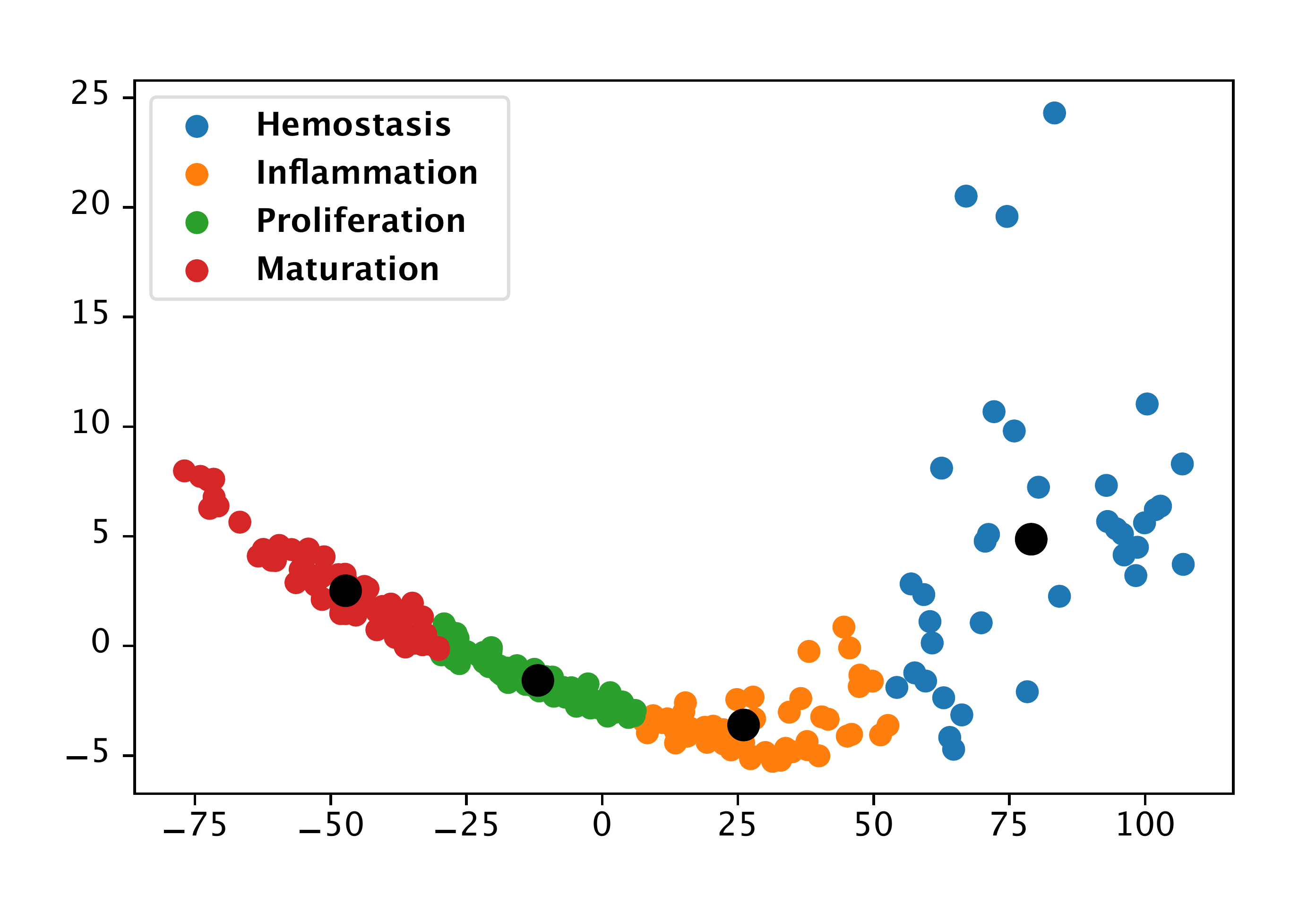}
    \caption{K-means ($k = 4$) cluster visualization with centroids using 2-dimensional PCA.}
    \label{fig:pca}
\end{figure}

We performed two-dimensional PCA in order to visualize the clusters along with their centroids. We can observe in Fig. \ref{fig:pca} that our clusters are mostly close together, especially in the later stages. Given how wound progression is slowly evolving and the time-step between observations is equal to 1 day, this closeness and some overlap are expected. We can also observe how the hemostasis cluster contains the highest spread. This is likely because this cluster mostly contains day 0 images where the wound bed is uncovered and artificially illuminated and day 1 images where the wound bed is no longer artificially illuminated and covered with a protective plastic film, causing some distance between the embeddings. Due to the noisy nature of the dataset, some spread is also expected.

\begin{figure}[t!]
    \centering
    \includegraphics[width=0.75\textwidth]{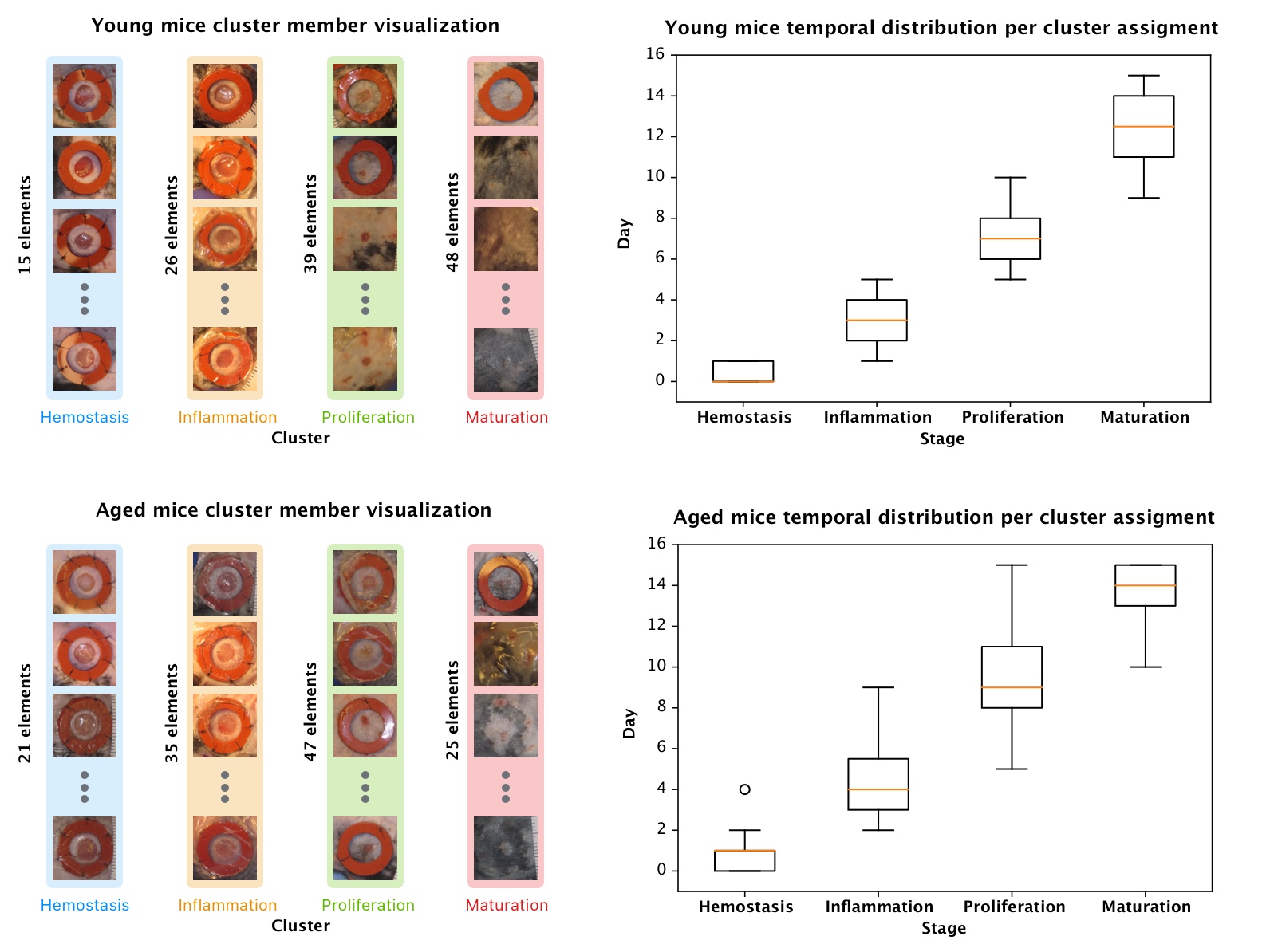}
    \caption{\textbf{(left)} Cluster member examples split between young and aged mice and \textbf{(right)} Temporal distribution box plots. Note: here we show the original non-cropped images to verify the medical red rings (splints) have not biased the cluster members.} We align the cluster groups with a heal-stage according to two metrics: visual appearances and wound age distribution metrics.
    \label{fig:stage}
\end{figure}

To further evaluate the quality of the clusters, we looked into the images in each cluster (for young and aged mice separately). The median wound age (in days) and first and third quartile of wound age are calculated for each cluster. As can be seen from Fig. \ref{fig:stage}, images in each of the clusters contain visual cues for hemostasis (fresh and clearly defined wound-edge), inflammation (swelling of the wound edge, wet and shiny appearance), proliferation (dry and matte appearance with rough and uneven texture), and maturation (complete or near-complete wound closure). In addition, there is a clear temporal progression in the median wound day age that also differentiates our clusters. It would be interesting to note that a differing wound healing rate between young and aged mice is encoded and shown in the median, first, and third quantiles of each cohort. Previous work on this dataset \cite{carrion2021automatic} also found a similar divergence in terms of wound-size progression between young and aged mice, aligning with these results. These two aspects are used in relating each cluster to a matching heal-stage. To validate these assignments, a blind survey of 5 human non-experts was performed where we found a top-1 agreement of over 80\% between the self-supervised and human labels. Table \ref{tab:table1} outlines detailed statistics on cluster temporal distribution and Table \ref{tab:table2} describes human annotator agreement per datasplit.

\begin{table}[h!]
\centering
\caption{Cluster group wound-age (experimental days 0 - 15) statistics.}
\resizebox{0.45\textwidth}{!}{
\begin{tabular}{||c|c|c|c|c|c||}
    \hline
    Age Group & Cluster & Q1 & Median(Q2) & Q3 & Average\\
    \hline
    \multirow{4}{*}{Young}&Cluster 1 & 0.0 & 0.0 & 1.0 & 0.5\\
    & Cluster 2 & 2.0 & 3.0 & 4.0 & 3.0 \\
    & Cluster 3 & 6.0 & 7.0 & 8.0 & 7.1\\
    & Cluster 4 & 11.0 & 12.5 & 14.0 & 12.5\\
    \hline
    \multirow{4}{*}{Aged}&Cluster 1 & 0.0 & 1.0 & 1.0 & 1.0\\
    &Cluster 2 & 3.0 & 4.0 & 5.5 & 4.5\\
    & Cluster 3 & 8.0 & 9.0 & 11.0 & 9.6\\
    & Cluster 4 & 13.0 & 14.0 & 15.0 & 13.4\\
    \hline
\end{tabular}
}
\label{tab:table1}
\vspace{-4mm}
\end{table}

\subsection{Wound Stage Classification}
The wound stage classification model is trained until convergence. While this task contains fewer training samples ($15\times$ less), the loss curves show the network is not overfitting, achieves relatively high accuracy on the validation set, and can predict the heal-stage of training, validation, and test sets (Table \ref{tab:table3}). The accuracy for the wound stage classification task is 95.8\%, 87.5\%, and 90.6\% on training, validation, and test sets, respectively. Finally, we trained an identical architecture as a straight-forward CNN Classifier (without the pre-text task) utilizing the collected human survey labels in order to compare performance. Under these conditions the network achieved a lower accuracy of 83.3\%, 81.3\%, and 78.1\% on training, validation, and test sets, respectively. Table \ref{tab:table3} summarizes the loss and accuracy scores described above.

\begin{table}[h]
\centering
\caption{Human annotator vs. automatic pseudo-label agreement.}
\resizebox{0.25\textwidth}{!}{
\begin{tabular}{||c|c||}
    \hline
    Data Split & Agreement (\%)\\
    \hline
    Train & 80.7 \\
    \hline
    Validation & 71.9 \\
    \hline
    Test & 87.5 \\
    \hline
    Overall & 80.5 \\
    \hline
\end{tabular}
}
\label{tab:table2}
\vspace{-8mm}
\end{table}

\begin{table}[h]
\centering
\caption{Performance metrics across both temporal coherency and wound stage classification tasks for training, validation, and test sets.}
\resizebox{0.6\textwidth}{!}{
\begin{tabular}{||c|c|c|c||}
    \hline
    Task & Train Acc. & Val. Acc. & Test Acc.\\
    \hline
    Supervised Classifier (baseline) & 83.3\% & 81.3\% & 78.1\% \\
    \hline
    Temporal Coherency (pre-text) & 97.2\% & 95.0\% & 97.7\% \\
    Wound Stage Classification (downstream) & \textbf{95.8\%} & \textbf{87.5\%} & \textbf{90.6\%} \\
    \hline
\end{tabular}
}
\label{tab:table3}
\vspace{-4mm}
\end{table}
\section{Conclusion}
HealNet is a self-supervised learning scheme for automatic wound stage discovery and classification. It forms a novel combination of pre-text temporal feature extraction, k-means clustering, and pseudo-label re-tasking and fine-tuning. The results demonstrate the effectiveness and performance of the presented method with zero human-assisted data annotation. Considering the cost-effectiveness of the approach compared to medical experts' continuous diagnosis, the developed pipeline is a promising tool for augmenting, accelerating, and improving traditional human diagnosis. Given that there exists a high chance for expert disagreement in the medical field, a self-supervised approach could aid in decision-making or in compiling a true `golden-master' list of visual indicators relating to the healing process, the healing stages and the temporal patterns between them. The proposed HealNet framework can also be used by the scientific community to discover obscure and unseen patterns in their unlabeled data at low cost.
\bibliographystyle{IEEEtran}
\bibliography{ref}

\begin{thebibliography}{10}
\providecommand{\url}[1]{#1}
\csname url@samestyle\endcsname
\providecommand{\newblock}{\relax}
\providecommand{\bibinfo}[2]{#2}
\providecommand{\BIBentrySTDinterwordspacing}{\spaceskip=0pt\relax}
\providecommand{\BIBentryALTinterwordstretchfactor}{4}
\providecommand{\BIBentryALTinterwordspacing}{\spaceskip=\fontdimen2\font plus
\BIBentryALTinterwordstretchfactor\fontdimen3\font minus
  \fontdimen4\font\relax}
\providecommand{\BIBforeignlanguage}[2]{{%
\expandafter\ifx\csname l@#1\endcsname\relax
\typeout{** WARNING: IEEEtran.bst: No hyphenation pattern has been}%
\typeout{** loaded for the language `#1'. Using the pattern for}%
\typeout{** the default language instead.}%
\else
\language=\csname l@#1\endcsname
\fi
#2}}
\providecommand{\BIBdecl}{\relax}
\BIBdecl

\bibitem{sen2009human}
C.~K. Sen, G.~M. Gordillo, S.~Roy, R.~Kirsner, L.~Lambert, T.~K. Hunt,
  F.~Gottrup, G.~C. Gurtner, and M.~T. Longaker, ``Human skin wounds: a major
  and snowballing threat to public health and the economy,'' \emph{Wound repair
  and regeneration}, vol.~17, no.~6, pp. 763--771, 2009.

\bibitem{steiner2020surgeries}
C.~A. Steiner, Z.~Karaca, B.~J. Moore, M.~C. Imshaug, and G.~Pickens,
  ``Surgeries in hospital-based ambulatory surgery and hospital inpatient
  settings, 2014: statistical brief\# 223,'' 2020.

\bibitem{guo2010factors}
S.~a. Guo and L.~A. DiPietro, ``Factors affecting wound healing,''
  \emph{Journal of dental research}, vol.~89, no.~3, pp. 219--229, 2010.

\bibitem{bagood2021re}
M.~D. Bagood, A.~C. Gallegos, A.~I.~M. Lopez, V.~X. Pham, D.~J. Yoon, D.~R.
  Fregoso, H.-y. Yang, W.~J. Murphy, and R.~R. Isseroff, ``Re-examining the
  paradigm of impaired healing in the aged murine excision wound model,''
  \emph{The Journal of investigative dermatology}, vol. 141, no.~4S, pp.
  1071--1075, 2021.

\bibitem{chino2020segmenting}
D.~Y. Chino, L.~C. Scabora, M.~T. Cazzolato, A.~E. Jorge, C.~Traina-Jr, and
  A.~J. Traina, ``Segmenting skin ulcers and measuring the wound area using
  deep convolutional networks,'' \emph{Computer methods and programs in
  biomedicine}, vol. 191, p. 105376, 2020.

\bibitem{oyibo2001comparison}
S.~O. Oyibo, E.~B. Jude, I.~Tarawneh, H.~C. Nguyen, L.~B. Harkless, and A.~J.
  Boulton, ``A comparison of two diabetic foot ulcer classification systems:
  the wagner and the university of texas wound classification systems,''
  \emph{Diabetes care}, vol.~24, no.~1, pp. 84--88, 2001.

\bibitem{ubbink2015predicting}
D.~T. Ubbink, R.~Lindeboom, A.~M. Eskes, H.~Brull, D.~A. Legemate, and
  H.~Vermeulen, ``Predicting complex acute wound healing in patients from a
  wound expertise centre registry: a prognostic study,'' \emph{International
  Wound Journal}, vol.~12, no.~5, pp. 531--536, 2015.

\bibitem{mirikharaji2021d}
Z.~Mirikharaji, K.~Abhishek, S.~Izadi, and G.~Hamarneh, ``D-lema: Deep learning
  ensembles from multiple annotations-application to skin lesion
  segmentation,'' in \emph{Proceedings of the IEEE/CVF Conference on Computer
  Vision and Pattern Recognition}, 2021, pp. 1837--1846.

\bibitem{larsson2016learning}
G.~Larsson, M.~Maire, and G.~Shakhnarovich, ``Learning representations for
  automatic colorization,'' in \emph{European conference on computer
  vision}.\hskip 1em plus 0.5em minus 0.4em\relax Springer, 2016, pp. 577--593.

\bibitem{zhang2016colorful}
R.~Zhang, P.~Isola, and A.~A. Efros, ``Colorful image colorization,'' in
  \emph{European conference on computer vision}.\hskip 1em plus 0.5em minus
  0.4em\relax Springer, 2016, pp. 649--666.

\bibitem{noroozi2016unsupervised}
M.~Noroozi and P.~Favaro, ``Unsupervised learning of visual representations by
  solving jigsaw puzzles,'' in \emph{European conference on computer
  vision}.\hskip 1em plus 0.5em minus 0.4em\relax Springer, 2016, pp. 69--84.

\bibitem{pathak2016context}
D.~Pathak, P.~Krahenbuhl, J.~Donahue, T.~Darrell, and A.~A. Efros, ``Context
  encoders: Feature learning by inpainting,'' in \emph{Proceedings of the IEEE
  conference on computer vision and pattern recognition}, 2016, pp. 2536--2544.

\bibitem{ledig2017photo}
C.~Ledig, L.~Theis, F.~Husz{\'a}r, J.~Caballero, A.~Cunningham, A.~Acosta,
  A.~Aitken, A.~Tejani, J.~Totz, Z.~Wang \emph{et~al.}, ``Photo-realistic
  single image super-resolution using a generative adversarial network,'' in
  \emph{Proceedings of the IEEE conference on computer vision and pattern
  recognition}, 2017, pp. 4681--4690.

\bibitem{gidaris2018unsupervised}
S.~Gidaris, P.~Singh, and N.~Komodakis, ``Unsupervised representation learning
  by predicting image rotations,'' \emph{arXiv preprint arXiv:1803.07728},
  2018.

\bibitem{chen2020improved}
X.~Chen, H.~Fan, R.~Girshick, and K.~He, ``Improved baselines with momentum
  contrastive learning,'' \emph{arXiv preprint arXiv:2003.04297}, 2020.

\bibitem{chen2020simple}
T.~Chen, S.~Kornblith, M.~Norouzi, and G.~Hinton, ``A simple framework for
  contrastive learning of visual representations,'' in \emph{International
  conference on machine learning}.\hskip 1em plus 0.5em minus 0.4em\relax PMLR,
  2020, pp. 1597--1607.

\bibitem{misra2016shuffle}
I.~Misra, C.~L. Zitnick, and M.~Hebert, ``Shuffle and learn: unsupervised
  learning using temporal order verification,'' in \emph{European Conference on
  Computer Vision}.\hskip 1em plus 0.5em minus 0.4em\relax Springer, 2016, pp.
  527--544.

\bibitem{yan2020clusterfit}
X.~Yan, I.~Misra, A.~Gupta, D.~Ghadiyaram, and D.~Mahajan, ``Clusterfit:
  Improving generalization of visual representations,'' in \emph{Proceedings of
  the IEEE/CVF Conference on Computer Vision and Pattern Recognition}, 2020,
  pp. 6509--6518.

\bibitem{carrion2021automatic}
H.~Carrión, M.~Jafari, M.~D. Bagood, H.-y. Yang, R.~R. Isseroff, and M.~Gomez,
  ``Automatic wound detection and size estimation using deep learning
  algorithm,'' \emph{PLOS Computational Biology}, 2022.

\bibitem{huang2017densely}
G.~Huang, Z.~Liu, L.~Van Der~Maaten, and K.~Q. Weinberger, ``Densely connected
  convolutional networks,'' in \emph{Proceedings of the IEEE conference on
  computer vision and pattern recognition}, 2017, pp. 4700--4708.

\bibitem{deng2009imagenet}
J.~Deng, W.~Dong, R.~Socher, L.-J. Li, K.~Li, and L.~Fei-Fei, ``Imagenet: A
  large-scale hierarchical image database,'' in \emph{2009 IEEE conference on
  computer vision and pattern recognition}.\hskip 1em plus 0.5em minus
  0.4em\relax Ieee, 2009, pp. 248--255.

\bibitem{bromley1993signature}
J.~Bromley, I.~Guyon, Y.~LeCun, E.~S{\"a}ckinger, and R.~Shah, ``Signature
  verification using a" siamese" time delay neural network,'' \emph{Advances in
  neural information processing systems}, vol.~6, 1993.

\bibitem{kingma2014adam}
D.~P. Kingma and J.~Ba, ``Adam: A method for stochastic optimization,''
  \emph{arXiv preprint arXiv:1412.6980}, 2014.

\end{thebibliography}

\end{document}


\title{Supplementary Materials for HealNet - Self-Supervised Acute Wound Heal-Stage Classification}
%
%
\author{Anonymous}
\authorrunning{Anonymous et al.}
%
\institute{Anonymous Organization \\
\email{**@****.***}} 
%
\maketitle              
%

%
%
%
\section{Software and Hardware}

\subsection{Setup}

All code involving this study was written in Python v3.7.12. The packages used for training and inference were the latest available stable versions of deep learning frameworks Keras v2.8 and TensorFlow v2.8. Additional packages used for numerical processing, plotting and visualizations were Pandas v1.3.5, NumPy v1.21.5, sklearn v1.0.2, PIL 7.1.2 and Matplotlib 3.5.1. The hardware used was a Google Colab instance running a quad-core Intel Xeon CPU at 2.3 GHz with 26GB of RAM and a Tesla P100 GPU with 16GB of vRAM.

\subsection{Performance}

The average run-time for pre-text training was about 10 minutes or ~25 seconds per epoch ($\sim$125ms/step) over 25 epochs. The average run-time for the downstream task was about a minute or ~2.5s per epoch ($\sim$150ms/step) over 25 epochs, this faster train-time is due to the smaller downstream dataset. Memory footprint did not exceed 16GB on either system memory or GPU memory after multiple runs. Inference time is equal to about 17 images per second.

\section{Additional Hyper-parameters}

For pre-text training, additional hyper-parameters used were 16 batch size, 25 epochs. These were selected as they allowed the model to converge relatively quickly with no over-fitting. For downstream training, additional hyper-parameters used were 12 steps per epoch, 2 validation steps per epoch and 25 epochs. These were selected as our training dataset contained 12 wound series and our validation dataset contained 2 wound series. Both steps include a 30\% dropout layer previous to the final fully connected layer, this value was chosen as it prevented over-fitting while allowing the model to converge.

\section{Full-width Figures}

This section contains all figures from the main text at full width to aid in viewing.

\begin{figure}[t]
    \centering
    \includegraphics[width=\textwidth]{Figures/Fig 1.pdf}
    \caption{\textbf{(Step 1)} The trained temporal encoder is used to generate 16-dimensional feature vectors. We cluster these feature vectors using k-means with $k=4$. \textbf{(Step 2)} The new pseudo-labels (based on clusters) are used to assign each image a heal-stage class. The temporal encoder is then re-tasked and fine-tuned using pseudo-labels, predicting the wound stage.}
    \label{fig:clustering}
\end{figure}

\begin{figure*}[t!]
    \centering
    \includegraphics[width=\textwidth]{Figures/Fig 2.pdf}
    \caption{\textbf{(left side)} Due to the nature of wound healing, images capturing progression naturally contain a temporal structure. We generate image pairs in the forwards (positive) and backward (negative) temporal directions; this serves as the input for our temporal encoder network. \textbf{(right side)} Randomly sampled image pairs.
}
    \label{fig:data}
\end{figure*}

\begin{figure*}[t!]
    \centering
    \includegraphics[width=\textwidth]{Figures/Fig 3.pdf}
    \caption{Positive and negative wound pairs train the temporal encoder. The encoder forms a Siamese tuple configuration with shared parameters; each branch outputs a 16-dimensional feature vector. The feature vectors are concatenated and passed through a fully-connected binary classifier, outputting the input pair temporal validity.}
    \label{fig:temporal-encoder}
\end{figure*}

\begin{figure}[h]
    \centering
    \includegraphics[width=\textwidth]{Figures/Fig 4.pdf}
    \caption{Training and validation loss and accuracy for the task of predicting positive or negative temporal pairs.}
    \label{fig:training-curves}
\end{figure}

\begin{figure}[h!]
    \centering
    \includegraphics[width=\textwidth]{Figures/Fig 5.pdf}
    \caption{K-means ($k = 4$) cluster visualization with centroids using 2-dimensional PCA.}
    \label{fig:pca}
\end{figure}

\begin{figure}[t!]
    \centering
    \includegraphics[width=\textwidth]{Figures/Fig 6.pdf}
    \caption{\textbf{(left)} Cluster member examples split between young and aged mice and \textbf{(right)} Temporal distribution box plots. We align the cluster groups with a heal-stage according to two metrics: visual appearances and wound age distribution metrics.}
    \label{fig:stage}
\end{figure}

\begin{figure}[h]
    \centering
    \includegraphics[width=\textwidth]{Figures/Fig 7.pdf}
    \caption{Training and validation loss and accuracy for the task of predicting heal-stage.}
    \label{fig:classification-results}
\end{figure}